\documentclass[runningheads]{llncs}
\usepackage[T1]{fontenc}
\usepackage{comment}

\usepackage{graphicx}
\usepackage{todonotes}
\usepackage[hidelinks]{hyperref}

\begin{document}
\title{Can Legislation Be Made Machine-Readable in PROLEG?}
\subtitle{An Investigation of GDPR Article 6}
%
%
\author{May Myo Zin\inst{1}\orcidID{0000-0003-1315-7704} \and
Sabine Wehnert\inst{2}\orcidID{0000-0002-5290-0321}
\and Yuntao Kong \inst{1}\orcidID{0009-0001-2089-2363}
\and Ha Thanh Nguyen\inst{1,3}\orcidID{0000-0003-2794-7010}
\and Wachara Fungwacharakorn\inst{1} \orcidID{0000-0001-9294-3118}
\and Jieying Xue \inst{1}\orcidID{0009-0000-8070-6609}
\and Michał Araszkiewicz \inst{4}\orcidID{0000-0003-2524-3976}
\and Randy Goebel\inst{5}\orcidID{0000-0002-0739-2946} 
\and Ken Satoh \inst{1}\orcidID{0000-0002-9309-4602} 
\and Nguyen Le Minh \inst{6}\orcidID{0000-0002-2265-1010} }
\authorrunning{Zin et al.}
%
\institute{Center for Juris-Informatics, ROIS-DS, Tokyo, Japan \and Ruhr-University Bochum, RC-Trust, Bochum, Germany 
\and Research and Development Center for Large Language Models, NII, Tokyo, Japan
\and Uniwersytet Jagielloński w Krakowie: Kraków, Poland
\and University of Alberta: Edmonton, Alberta, Canada
\and Japan Advanced Institute of Science and Technology, Ishikawa, Japan }

%
\maketitle              
\begin{abstract}
The anticipated positive social impact of regulatory processes requires both the accuracy and efficiency of their application. Modern artificial intelligence technologies, including natural language processing and machine-assisted reasoning, hold great promise for addressing this challenge. We present a framework to address the challenge of tools for regulatory application, based on current state-of-the-art (SOTA) methods for natural language processing (large language models or LLMs) and formalization of legal reasoning (the legal representation system PROLEG). As an example, we focus on Article 6 of the European General Data Protection Regulation (GDPR). In our framework, a single LLM prompt simultaneously transforms legal text into if-then rules and a corresponding PROLEG encoding, which are then validated and refined by legal domain experts. The final output is an executable PROLEG program that can produce human-readable explanations for instances of GDPR decisions. We describe processes to support the end-to-end transformation of a segment of a regulatory document (Article 6 from GDPR), including the prompting frame to guide an LLM to ``compile'' natural language text to if-then rules, then to further ``compile'' the vetted if-then rules to PROLEG.  Finally, we produce an instance that shows the PROLEG execution. We conclude by summarizing the value of this approach and note observed limitations with suggestions to further develop such technologies for capturing and deploying regulatory frameworks.


\end{abstract}
%
%
\section{Introduction}
Modern legal and regulatory texts, such as the EU General Data Protection Regulation (GDPR), are written for human interpretation. Their open-textured language, cross-references, and exceptions make them difficult to operationalize in formal models, let alone in software \cite{araszkiewiczpleszka2015}. As organizations seek to automate compliance at scale, this human-centric drafting becomes a bottleneck: it impedes consistent interpretation, slows audits, and raises the cost of demonstrating conformity. Converting law into a structured, machine-readable form is therefore a prerequisite for AI-assisted legal analysis, explainable decision support, and robust, auditable compliance automation.

\begin{figure}
  \centering
  \includegraphics[width=.9\linewidth]{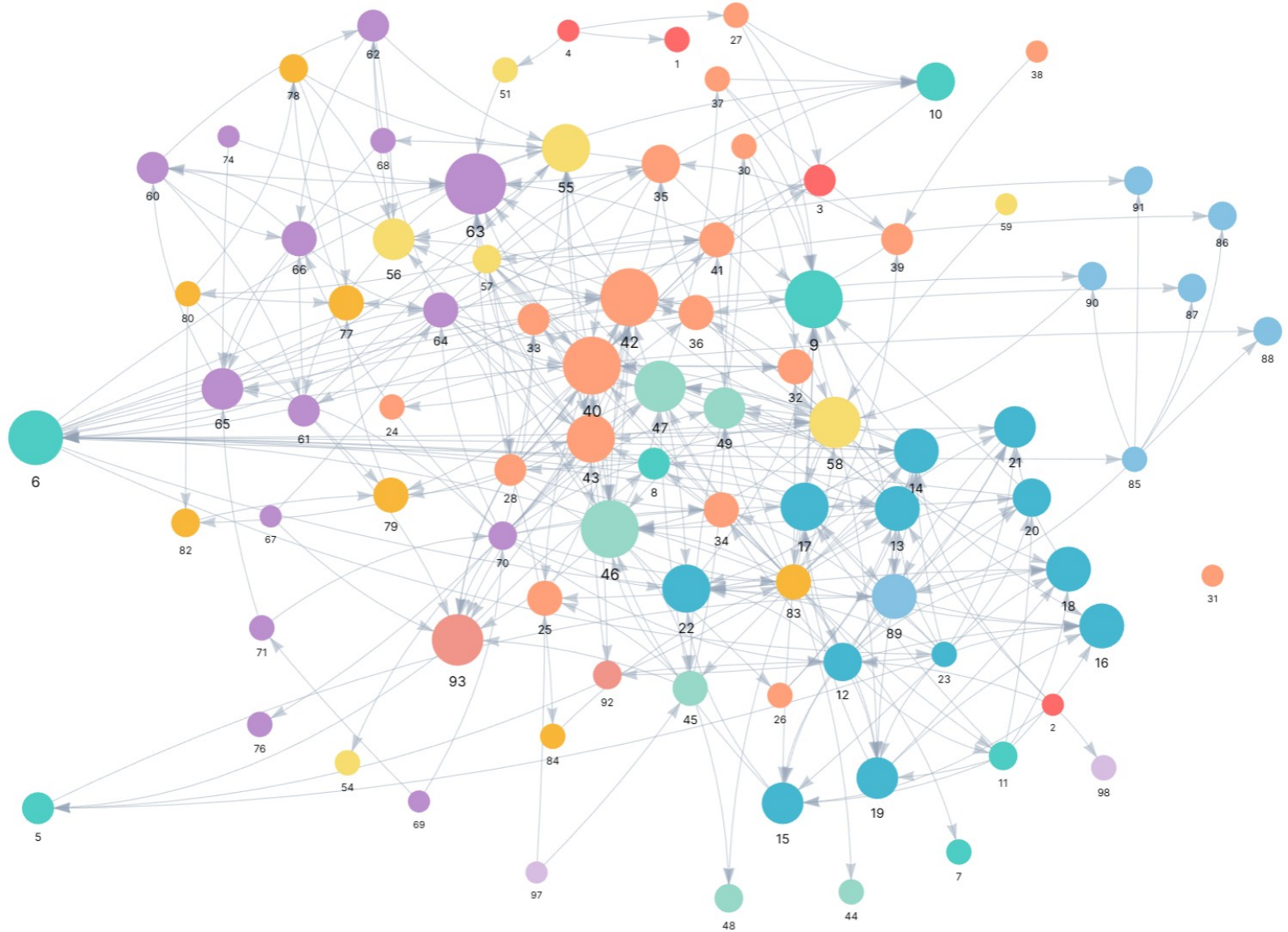}
  \caption{Network of explicit cross-references among GDPR Articles.}
  \label{fig:gdpr-mentions}
\end{figure}

This paper presents a human-in-the-loop workflow that transforms natural-language provisions into executable rules suitable for automatic reasoning and compliance checking. We focus on GDPR Article 6 (lawfulness of processing) as a representative, high-impact provision whose nuanced conditions and exceptions exemplify the challenges of formalization. Our approach combines large language models (LLMs) for scalable drafting with expert validation to ensure legal fidelity, and targets a logic-based formalism (PROLEG) to support sound, \emph{inspectable} inference. 
As illustrated by Figure~\ref{fig:gdpr-mentions}, the GDPR exhibits a dense network of inter-article references. This figure visualizes only explicit connections, yet this interconnectedness implies that analyzing Article 6 will also require consideration of related provisions.

The PROLEG knowledge representation language \cite{Satoh2023} was developed to facilitate interactions between lawyers and legal reasoning systems. While it does not address all challenges---such as limited expressiveness for certain legal concepts and the ambiguity of legal texts---it does provide a minimal yet sufficient language for reasoning, thus enabling lawyers to understand system behavior. 

Our approach pairs a single, fixed composite prompt—designed to produce both if–then rules and an initial PROLEG encoding from the article text—with expert review that ensures doctrinal fidelity and preserves the semantic structure of the provision. By grounding LLM-generated rule candidates in expert validation and an executable formalism, the framework supports transparent reasoning and traceability from formal rules back to the authoritative text. 

The contributions of this work are threefold:
\begin{itemize}
    \item A practical workflow that couples LLM-based rule generation with expert review to produce PROLEG-executable legal rules;
    
    \item A qualitative evaluation of the doctrinal fidelity of if–then and PROLEG representations produced by a fixed composite prompt; and
    \item A curated rule set and test cases for GDPR Article 6, together with a demonstration of end-to-end reasoning behavior, including failure modes.
\end{itemize}

We conclude by examining the limitations of the initial composite prompt approach, including residual ambiguity and challenges associated with transferring the method to new legal domains, and by outlining how the pipeline may generalize to other GDPR provisions and regulatory frameworks.

\section{Background}
Here we present background on the PROLEG logical framework and collect related work on formal representations of legal knowledge, spanning logic- and rule-based systems, ontology and knowledge-graph approaches, and machine-readable standards for normative texts.

\subsection{PROLEG}

PROLEG (short for PROlog-based LEGal reasoning support system) \cite{satoh2010proleg} is a logic programming framework developed to model and support legal reasoning. It is based on the Prolog language but is specifically designed to represent legal rules, exceptions, and facts in a structured and executable form.

The system was originally proposed to formalize reasoning under the Presupposed Ultimate Fact Theory (JUF theory) in Japanese jurisprudence. Unlike standard logic programming, PROLEG introduces constructs that capture the nuances of legal argumentation, such as exceptions, burden of proof, and rule hierarchies, which are central to legal decision-making.

In PROLEG, laws are encoded as logical rules (similar to Horn clauses), while the facts of a case are represented separately. The reasoning process then derives conclusions by applying the rules to the facts, taking into account exceptions and conflicting interpretations. This allows PROLEG to simulate the reasoning process of courts and legal practitioners in a transparent and explainable way.

PROLEG has been used primarily in research on computational legal reasoning and AI \& Law, particularly for analyzing statutory interpretation and case reasoning in civil law systems. It provides a visual reasoning trace, ensuring that the resulting conclusions are both interpretable and reproducible, thereby enhancing the transparency of the reasoning process. Although it is not a general-purpose programming language, it serves as an important bridge between symbolic AI and legal knowledge representation, contributing to the broader field of legal informatics. For illustration purposes, we consider the following use case for formalization.


\textit{A financial institution collected an extensive set of personal data from individuals on the basis of their consent. The institution processed the data, inter alia, for the purpose of marketing. Later, one of these individuals withdrew the consent and asked the institution to stop using the data. Despite this withdrawal, the institution intended to continue processing the data.}

\begin{figure}[htbp]
    \centering
    \includegraphics[width=\textwidth]{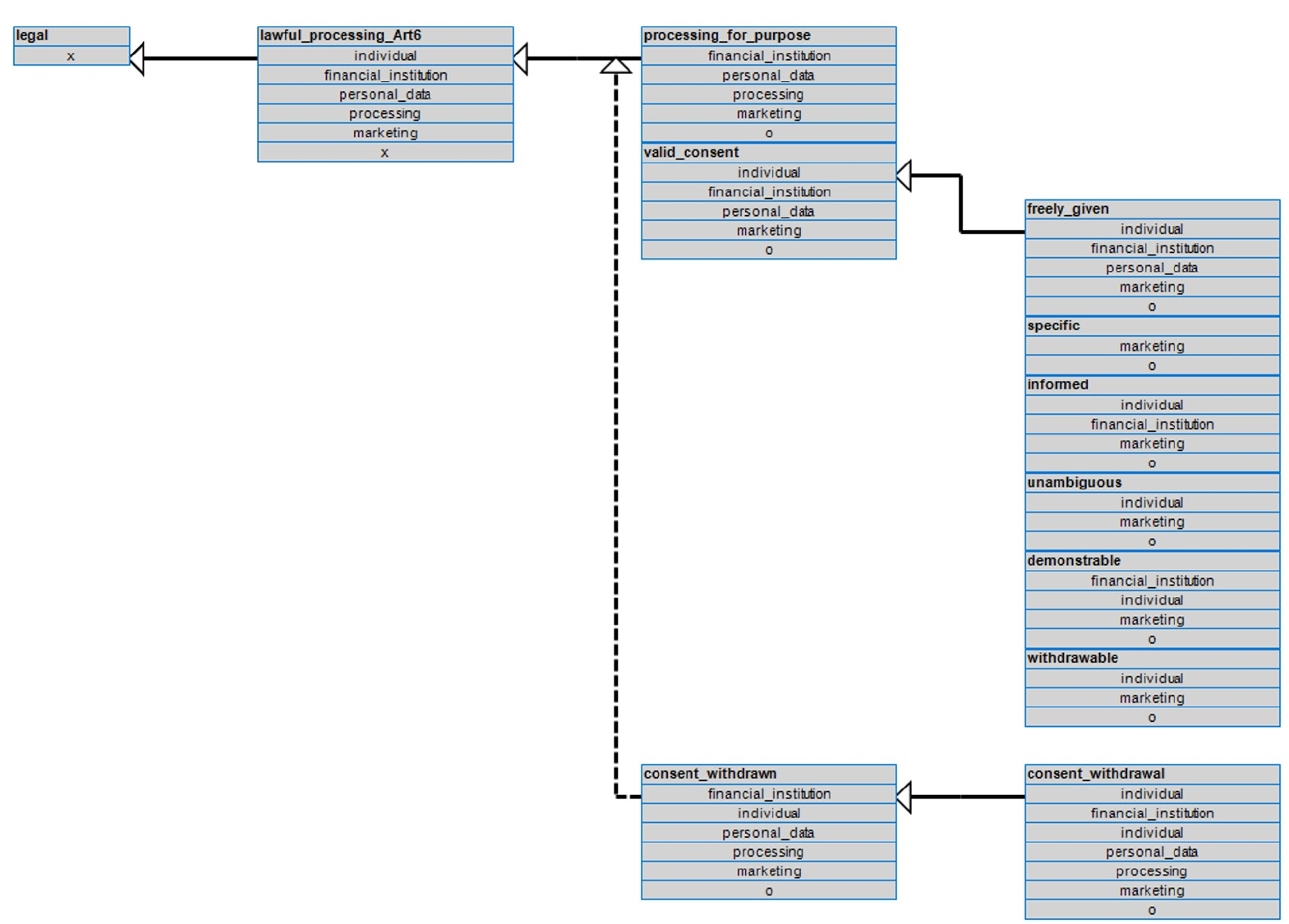}
    \caption{Example of PROLEG block diagram.}
    \label{proleg_output}
\end{figure}

This use case has been transformed into PROLEG and concerns GDPR Article 6. The output result is shown in the form of a block diagram in Figure~\ref{proleg_output}. A bottom item of each block expresses the result of evaluation of conclusions or
conditions (o: success, x: fail). A solid arrow between blocks expresses the conclusion-condition relation for a general rule, while a dotted arrow shows the exception relation of the conclusion of a general rule. The reasoning concludes with an overall failure mark (x), indicating that continued processing after consent withdrawal is unlawful.

\subsection{Logic-Based and Rule-Based Approaches}

Early efforts focused on using logic-based systems to model legal rules and reasoning. Logic programming, particularly Prolog, has been used as a foundation for these systems. A prominent example is the PROLEG system (PROlog-based LEGal reasoning support system), developed to implement the Presupposed Ultimate Fact Theory of Japanese civil law \cite{satoh2010proleg}. This system demonstrates how legal rules can be represented as Horn clauses and then processed using Prolog technology to support legal reasoning, particularly in contexts with incomplete information.

The field has also explored non-monotonic logics such as defeasible logic and deontic logic to capture further nuances of legal reasoning, such as exceptions to rules and concepts of obligation, permission, and prohibition. The pioneering work of Jones et al. \cite{jones1992deontic} explored the use of deontic logic to represent legal norms. Hybrid approaches have been proposed, such as that of Dragoni et al. \cite{dragoni2016combining}, which combines different Natural Language Processing (NLP) techniques, including frame-based, dependency-based, and logic-based extraction, to extract machine-readable rules from legal texts. This hybrid approach demonstrates the necessity of integrating multiple methods to address the complexity of legal language. Similarly, work on normative texts within the Norme in Rete (NIR) project introduces machine-learning-based provision classifiers and rule-based argument extractors that operate at the level of individual provisions, automatically identifying provision types and their arguments to enrich legal documents with structured semantic metadata \cite{biagioli2005automatic}. In our work, we likewise focus on the provision level, but our intention is to directly work on the provision text itself as the primary input to the reasoning pipeline, rather than assuming a separate, pre-annotated semantic layer.

Building upon PROLEG, Nguyen et al. \cite{nguyen2022interactive} developed an interactive natural language interface to make the system more accessible to legal practitioners unfamiliar with PROLEG. The system consists of three main modules: a natural language perceiver, a PROLEG reasoner, and an inference explainer, thereby bridging the gap between formal logic representations and natural language input from lawyers.
Subsequently, Nguyen et al. \cite{Nguyen2025DataAugmented} proposed a pipeline for a Deep PROLEG system, which addressed scalability issues when synthesizing artificial data of legal cases in new domain adaptations.

\subsection{Ontology-Based and Knowledge Graph Approaches}

Another important research direction involves using ontologies to formally represent legal knowledge. Ontologies provide a shareable vocabulary and structure to describe concepts and relationships in the legal domain. Legal ontology engineering methodologies, as detailed by Casellas et al. \cite{casellas2011legal}, provide systematic approaches to building and maintaining these knowledge representations. Relevant in our context is the work by Palmirani et al. \cite{palmirani2018pronto} who introduce PrOnto, a modular GDPR-oriented privacy ontology developed with the MeLOn methodology that formally models data, actors, processing workflows, purposes, legal bases, and deontic norms to support automated legal reasoning and compliance checking using semantic web technologies. Leone et al. \cite{leone2020taking} systematically compare and classify a set of existing legal ontologies across general, modeling, and semantic dimensions. They highlight strengths, weaknesses, and reuse potential in order to guide users in selecting and extending suitable ontologies for legal knowledge representation. More recently, knowledge graphs have emerged as an extension of ontology-based methods, which are claimed to enable more flexible and scalable representation of legal entities and relationships. 
Servantez et al. \cite{servantez2023computable} further advance this line of work by introducing a graph-based representation of contract obligations (Obligation Logic Graphs) that supports automated conversion of natural-language contracts into code; this further bridges the gap between legal ontologies and executable contract logic. In contrast, our work focuses on natural-language legal texts rather than formal ontologies, to leverage their linguistic richness and contextual depth to better capture meaning and reasoning in real-world legal documents.

\subsection{Machine-Readable Standards}

To ensure interoperability and widespread adoption, standardization of machine-readable formats for legal texts is important. As this standardization develops, several approaches have emerged. For example, \textbf{Akoma Ntoso} (Architecture for Knowledge-Oriented Management of African Normative Texts using Open Standards and Ontologies) is a prominent XML standard for representing parliamentary, legislative, and judicial documents \cite{palmirani2011legislative}. It provides a rich vocabulary for marking up the structure and semantics of legal texts.

Building on this foundation, \textbf{LegalRuleML} has been developed as another standard for representing legal rules in a machine-readable format. Lam et al. \cite{lam2019enabling} demonstrated how LegalRuleML can be used to enable reasoning by transforming represented rules into modal defeasible logic, thereby bridging the gap between rule representation and automated reasoning. More recently, new semantic formats such as \textbf{X2RL} (eXplainable, eXtractable, Rule-like Language) have been proposed by McLaughlin et al. \cite{mclaughlin2021drafting} to augment regulatory documents with rich metadata fields, and focus not only on structure but also on content and meaning, which aims to reduce regulatory management and compliance costs.

Logic-based systems make norms executable but need manual rule crafting and expert use; ontology and graph methods model structure but not executable reasoning; and machine-readable standards ensure interoperability yet stop before inference. Our work bridges these by turning natural-language provisions directly into validated, PROLEG-executable rules through an LLM-expert workflow. This yields traceable, auditable, and explainable reasoning while remaining lightweight and compatible with existing ontologies and standards.

\section{Methodology}
In this section, we describe our methodology and detail the prompt design.

\subsection{Overview of the Approach}

Our goal is to examine how a single composite LLM prompt can support the transformation of legal text into PROLEG-executable rules. To ensure a controlled evaluation, the study uses one fixed prompt, applied once to each target provision. The outputs of this prompt, namely a set of if–then rules and an initial PROLEG encoding, form the basis for expert analysis.

\begin{figure}[htpb]
    \centering
\includegraphics[width=\linewidth]{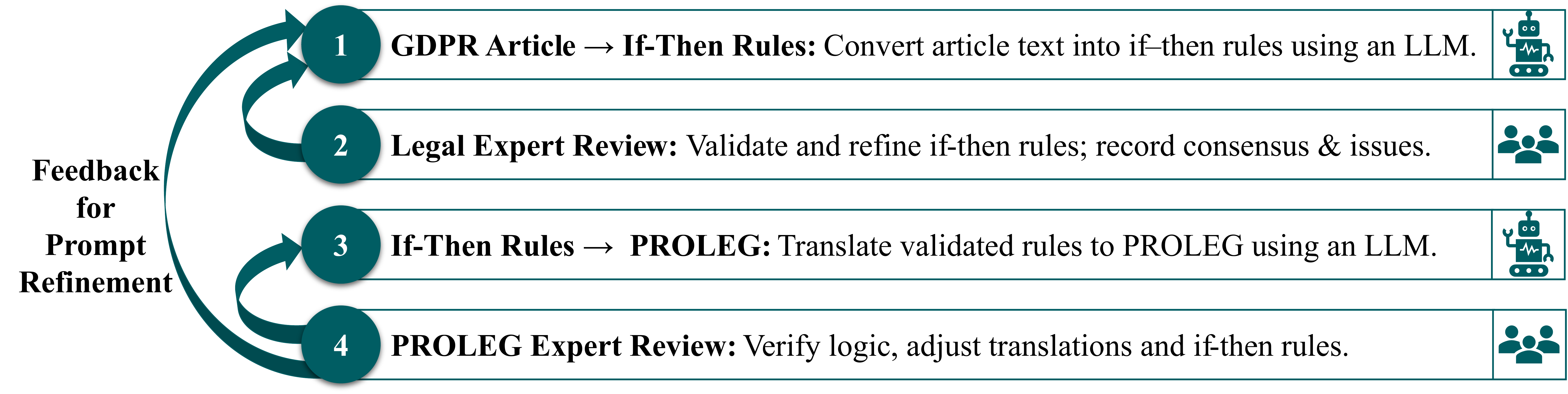}
    \caption{Overview of the  GDPR to PROLEG conversion process.}
    \label{fig:process}
\end{figure}

Figure \ref{fig:process} presents the conceptual workflow, which consists of four stages involving both automated generation and expert review. Although the figure separates these stages for clarity, in this study both the if–then rules and the initial PROLEG encoding are produced by a single composite prompt.

The workflow proceeds as follows:

\begin{itemize}
    \item [1.] \textbf{Generation of If–Then Rules and initial PROLEG Encoding:} The composite prompt instructs the LLM to interpret the selected GDPR provision, extract its conditions and exceptions, express them as structured if–then rules, and generate a corresponding PROLEG-style representation. These two representations serve as preliminary formalizations of the provision.
    \item[2.] \textbf{Legal Expert Review:} Legal experts examine the if–then rules to assess whether the extracted structure accurately reflects the provision’s normative content. Their review focuses on the allocation of information across main rules and sub-rules, the handling of exceptions, and the preservation of legally significant distinctions.
    \item[3.] \textbf{PROLEG Expert Review:} Specialists in PROLEG evaluate the LLM-generated formalization for syntactic correctness and intended semantics. They identify issues such as predicate mismatches, logical gaps, or distortions introduced by the model’s interpretation of the text.
    \item[4.] \textbf{Case Creation and Execution:} A case-creation team constructs representative factual scenarios, defines the necessary fact schema, and encodes these facts. The validated rule set and encoded facts are then executed in a PROLEG environment, producing reasoning traces that reveal how the formalized rules behave in concrete cases.
\end{itemize}
This design allows us to isolate and analyze the characteristic patterns of a fixed-prompt approach, both its strengths and its limitations, while keeping LLM generation, expert evaluation, and PROLEG execution clearly separated. We systematically record recurring issues in the structure or wording of the rules and in their PROLOG counterparts as part of our evaluation.

\subsection{Prompt Design}



The interpretation of the GDPR is inherently complex due to the interdependencies among its provisions, which frequently span multiple Articles and their corresponding Recitals; therefore, a comprehensive and systematic interpretative approach must consider these cross-references and contextual relationships. 
This study employs a custom version of ChatGPT-5 in which the Chain-of-Instructions (COI) prompt \cite{zin2024leveraging} is embedded as an internal reasoning framework; this guides the model through each stage of legal interpretation and rule formalization. 
The prompt operates through five internal instruction stages. The first two stages instruct the model to identify all relevant Recitals and cross-referenced Articles that provide interpretive or procedural context for the selected GDPR provision. This ensures that the subsequent interpretation and formalization processes are contextually grounded and legally coherent. The third stage directs the synthesis of these interpretive materials into a unified, detailed legal rule that explicitly incorporates all dependencies and interpretive nuances. This stage is critical for maintaining normative fidelity, as omissions or simplifications could compromise the representational validity of the machine-readable rule.
In the fourth stage, the synthesized rule is transformed into a normalized if-then structure. This logical framing not only enforces syntactic clarity but also aligns the rule with the representational requirements of logic-based systems. It explicitly separates conclusions, conditions, and exceptions, thereby supporting transparency in both human and machine reasoning. The final stage involves generating a complete PROLOG program that includes entities, rules, exceptions, facts, and queries. Although our ultimate goal is to produce a PROLEG program, we initially instruct the model to generate the logic in PROLOG, since current LLMs are more familiar with it. This approach helps achieve higher accuracy, and we can then systematically convert the resulting PROLOG program into PROLEG. The complete prompt and the corresponding initial outputs generated for GDPR Article 6 (Lawfulness of Processing) are available in the GitHub repository\footnote{\url{https://github.com/JurisInformaticsCenter/GDPR-PROLEG-data}}.


\section{Results and Analysis}

The generated if-then rules were subject to qualitative evaluation performed by legal experts. This evaluation was based on the criterion of the adequacy of the generated rules to the structure and content of legal norms that lawyers derive from the GDPR provisions expressed in natural language. Importantly, our approach relied on the structural resemblance assumption, meaning that a representation of a rule based on one Article (or a part thereof) is preferred over complex representations derived from multiple Articles. If other parts of the normative material are relevant, e.g., for the interpretation of the generated rule, they should be presented in the form of accompanying sub-rules rather than as components of the general rule. This semantic layering of the generated rules was verified against professional doctrinal material \cite{litwinski2025praxis}.
Different types of infidelity may arise between the if-then rule and the original normative material, including confused or otherwise altered structure, as well as semantic modifications that may lead either to restriction or broadening of the target rule. One research question of the project is to investigate what types of inadequacies or other issues occur in the generated content, assuming the fixed prompting system defined above.
We did not attempt to develop a complete classification of possible inadequacy issues. At the outset, we adopted only a general distinction between structural and semantic issues. We decided to develop a typology of such issues using a bottom-up method, that is, by analysing specific outputs and clarifying the nature of the identified issues. The resulting catalogue of issue examples is presented in the following two subsections: structural issues (Section 4.1) and semantic issues (Section 4.2).

\subsection{Structural Issues}

\subsubsection{Questionable allocation of information to sub-rules:}

ChatGPT generated catalogues of sub-rules for the main rules, but in certain instances, it was unclear why one piece of information was included in the set of sub-rules while another, equally relevant, was not. For example, in the context of Art. 6.1(a) (consent), ChatGPT generated sub-rules explaining that consent is freely given (Art. 7.4) and that consent should be withdrawable (Art. 7.3), but then omitted the requirement that consent should be distinguishable from other statements and expressed in clear language (Art. 7.2). Similarly, it created exception-like sub-rules indicating parental authorization of a child’s consent (Art. 8) and explicit consent for the processing of special categories of data (Art. 9.1(a)), but not other instances of explicit consent (Art. 22(c), Art. 49.1(a)).
In another result, ChatGPT generated a sub-rule related to Arts. 13 and 14 (in representing Art. 6.1(c)), while (correctly) refraining from doing so in other cases.

\subsubsection{Questionable allocation of information to main rules:}

In some instances, ChatGPT enriched the catalogue of conditions for the main rule extracted from a specific Article by resorting to other Articles. For example, in the output for Art. 6.1(a), it included the demonstrability requirement (Art. 7.1) in the body of the rule, as opposed to the withdrawability requirement (Art. 7.3), which was placed in the sub-rules.
Similarly, in the generated main rule representing Art. 6.1(b) (contractual necessity), ChatGPT added a condition that processing should be compliant with the general principles outlined in Art. 5. However, this condition applies to any instance of processing under the GDPR and should therefore be included in the condition set of every rule concerning data processing. This was not done consistently. Our conclusion was that such results should be avoided for the sake of structural resemblance.


\subsubsection{Generating presupposed clauses:}

In some instances concerning the reconstruction of rules regarding data processing, ChatGPT reconstructed presupposed information (“data is processed”), as in the case of Art. 6.1(c), whereas in other instances it did not. Generally, this should be avoided because, if a rule specifies the conditions for the legality of data processing, its applicability already presupposes that data processing is occurring; thus, the information “data is processed” is already implicit in any condition that mentions processing.

\subsection{Semantic Issues}

\subsubsection{Predicate simplification:}

The generated sub-rule explaining the term “being freely given” simplified the condition excluding compliance with this requirement. The generated rule reads: “If consent is conditional upon services not necessary for contract performance, then it is not freely given.” However, such conditionality is not, in fact, a sufficient reason to conclude that consent is not freely given. In some cases, despite the presence of a conditional mechanism, consent may still be freely given. This is because the existence of such a mechanism is gradual rather than binary: the more intensively it is present in the analysed case, the stronger, ceteris paribus, the argument that consent might not have been freely given. Additionally, the presence of a conditional mechanism is only one among many reasons that may lead to non-compliance with the requirement: ChatGPT omitted the “inter alia” clause present in Art. 7.4.

\subsubsection{Questionable paraphrases or inference results as sub-rules:}

In some instances, instead of providing explanatory or definitional information in sub-rules, ChatGPT attempted to paraphrase the main rule or to draw an inference from it. This occurred in the context of Art. 6.1(b). The natural-language expression of the rule reads: “processing is necessary for the performance of a contract to which the data subject is party or in order to take steps at the request of the data subject prior to entering into a contract,” whereas the generated sub-rule stated: “If the purpose of processing is not necessary for performing or entering a contract, then processing is not lawful under Article 6(1)(b).”
Although this is, strictly speaking, a valid inference, it omits important information: processing of data prior to entering the contract must occur at the request of the data subject, not the controller. Consequently, if we intend to draw an inference from the main rule to a negative conclusion (i.e., specify when there is non-compliance with the Article), we should also state that even if such processing were necessary prior to entering the contract, it would still be non-compliant if initiated by the controller.

\subsubsection{Adding information not present in the relevant source text:}

In sub-rule 3 generated for Art. 6.1(c), ChatGPT stated: “If the legal obligation is derived from non-EU/non-Member State law without an EU legal mandate.” The phrase “without an EU legal mandate” does not appear in the source text. Art. 6.3, which clarifies the sources of relevant legal obligations, mentions only Union law or Member State law. Moreover, the term “mandate” is used in the GDPR in different contexts, referring to authorizing an entity to act on behalf of someone.

\subsubsection{Restricting the semantic scope of expressions:}

While generating a rule representing Art. 6.1(d), ChatGPT restricted the crucial expression “vital interest” to “vital interest essential for the life or physical integrity.” Although the resulting scope includes instructive examples, it omits a significant part of the original scope; for example, private property may also be classified as a “vital interest.”


\section{Methodological Limitations}

Although our goal was to explore an end-to-end transformation pipeline for converting natural-language legal provisions into executable PROLEG rules, we deliberately limited the study to a single custom LLM configuration and to GDPR Article 6. This constrained scope enabled us to isolate the effects of prompt design, model behavior, and expert feedback without introducing additional variability from cross-model differences or broader regulatory contexts. Accordingly, the present analysis is best interpreted as a proof-of-concept rather than a comprehensive empirical evaluation. Future work will extend the methodology to multiple LLMs, additional GDPR provisions, and larger, more heterogeneous test sets to systematically assess robustness and generalizability.

Within this constrained setting, the methodology demonstrates the feasibility of linking legal interpretation and logical formalization within a single prompt, but several limitations must be acknowledged. The current approach remains sensitive to model behavior, even when using the same prompt and identical input text. When applied repeatedly to the same Article, the number and ordering of extracted Recitals and cross-referenced Articles occasionally differ slightly across runs. Although the LLM consistently identifies the most important and directly relevant provisions, it occasionally omits provisions that are not explicitly connected but remain contextually essential for a complete legal interpretation. For example, when applied to Article 6(1)(a) (lawfulness of processing based on consent), the model successfully retrieved the directly related and most relevant provisions, such as Articles 4(11), 7, 8, and 9(2)(a), as well as key Recitals 32, 33, 42, and 43. However, it failed to include two important Articles, namely Article 49(1)(a) and (f), and Article 22(1) and (2), which provide necessary contextual and substantive links regarding data transfers based on explicit consent and the prohibition of automated individual decision-making, including profiling. These omissions suggest that, while the model performs effectively in identifying core definitional and consent-related provisions, it struggles to capture cross-contextual dependencies that are less syntactically but more semantically connected, thereby highlighting a limitation in its ability to model the broader legal relationships inherent in the GDPR framework. 


Moreover, the methodology was intentionally confined to a single prompt with an internal Chain-of-Instructions (CoI) executed by a single-model configuration. 
Although this design enabled an assessment of the model’s capacity for integrated, end-to-end legal text formalization, future work could explore multi-agent or modular architectures in which specialized sub-models independently address interpretive, logical, and representational tasks. Such distributed systems could enhance interpretability, mitigate reasoning drift, and support feedback loops between human experts and computational agents.
\section{Discussion}

The results of our qualitative evaluation demonstrate both the promise and the current limitations of using LLMs for the structured reconstruction of legal norms. On the one hand, the models are capable of producing rule-like representations that reflect, at least superficially, the architecture of statutory provisions. On the other hand, our analysis reveals systematic deviations from the expected structural and semantic fidelity. These deviations arise not merely at the level of technical inaccuracies but often concern deeper issues related to the allocation of information, the preservation of essential normative distinctions, and the interpretation of legal terminology. The structural issues show that the model lacks a stable internal criterion for distinguishing between the general rule and its sub-rules, leading to an inconsistent distribution of content across representational layers. Similarly, the semantic issues indicate a tendency toward overgeneralisation, oversimplification, or unwarranted inferences, which can narrow or distort the scope of the reconstructed norm. Taken together, these findings suggest that while current prompting methods can elicit rules resembling expert-generated representations, they do not guarantee their doctrinal adequacy. The observed patterns of inadequacy underscore the need for comparing different prompting strategies, model calibration, and perhaps hybrid systems that combine linguistic generation with domain-specific legal constraints.

\section{Conclusion and Future Work}
We have examined whether GDPR Article 6 can be rendered machine-readable in PROLEG through a human-in-the-loop workflow that combines large language models with expert review. We showed how a single chain of instructions prompt can guide the model from identifying relevant Articles and Recitals, through drafting if-then rules, to compiling executable PROLEG code and test cases in a demo system for compliance checks. Our qualitative evaluation indicates that this LLM-based compilation process can recover much of the structure of legal norms, while also revealing recurring structural and semantic issues, such as unstable allocation of content across main rules and sub-rules, reconstruction of presupposed clauses, and simplifications or additions that alter the scope of key predicates. We deliberately assumed this prompting basis as a static approach and observed what followed from that choice, but future work will improve the prompts, for example, by explicitly instructing the model not to paraphrase the wording of GDPR Articles and Recitals, because such paraphrasing introduced additional ambiguity. These findings confirm the value of PROLEG as an inspectable target formalism, highlighting the suitability of LLMs as generators of rule candidates. It further demonstrates that expert validation is essential for doctrinally reliable machine-readable legislation. Future work will refine prompt design in light of these insights, strengthen guidance for LLM-based rule and code generation, and extend the general approach beyond Article 6 to other parts of the GDPR and to further bodies of legislation.

    
\begin{credits}
\subsubsection{\ackname} This research was supported by ROIS-DS-JOINT (023RP2025), the “Strategic Research Projects” grant from ROIS (Research Organization of Information and Systems), the “R\&D Hub Aimed at Ensuring Transparency and Reliability of Generative AI Models” project of the Ministry of Education, Culture, Sports, Science and Technology, and JSPS KAKENHI Grant Numbers JP22H00543, JP25H00522,  JP25H01112, and JP25H01152.
\end{credits}

%
%
%

\vspace{2em}
\begingroup
\let\clearpage\relax   
\bibliographystyle{splncs04}
\bibliography{mybib}
\endgroup

\end{document}